\documentclass[11pt]{style/prism-princeton}

\usepackage{amsmath,amssymb,stmaryrd,mathtools}
\usepackage{xcolor}
\usepackage{xspace}
\usepackage{bbm}
\usepackage{graphicx}
\usepackage{subfig}
\usepackage{tcolorbox}
\usepackage{colortbl}
\usepackage{multirow}
\usepackage{booktabs}
\usepackage{wrapfig}

\definecolor{dark_red}{RGB}{122, 0, 0}
\definecolor{coral}{RGB}{255, 119, 94}
\definecolor{pink_orange}{RGB}{255, 72, 126}
\definecolor{vibrant_pink}{RGB}{255, 0, 104}
\definecolor{pink_pink}{RGB}{255, 37, 153}
\definecolor{wine}{RGB}{204, 0, 102}

\definecolor{light_orange}{RGB}{255, 198, 107}
\definecolor{orange(sae/ece)}{rgb}{1.0, 0.49, 0.0}
\definecolor{dark_orange}{RGB}{216,92,0}

\definecolor{org-purp-0}{RGB}{165, 76, 0}
\definecolor{org-purp-1}{RGB}{250, 130, 28}
\definecolor{org-purp-2}{RGB}{226, 89, 68}
\definecolor{org-purp-3}{RGB}{206, 92, 124}
\definecolor{org-purp-4}{RGB}{116, 80, 146}
\definecolor{org-purp-5}{RGB}{110, 78, 157}

\definecolor{teal(sae/ece)}{rgb}{0, 0.47, 0.52}
\definecolor{aqua}{RGB}{52,172,139}
\definecolor{dark_aqua}{RGB}{35,115,93}
\definecolor{dark_green}{RGB}{0, 92, 34}

\definecolor{grape}{RGB}{112,48,160}
\definecolor{purple}{rgb}{0.74, 0.65, 1.0}
\definecolor{dark_purple}{rgb}{0.58, 0.0, 0.82}
\definecolor{periwinkle}{RGB}{191, 140, 230}

\definecolor{light_gray}{rgb}{0.9, 0.9, 0.9}
\definecolor{medium_gray}{rgb}{0.6, 0.6, 0.6} 
\definecolor{dark_gray}{rgb}{0.2, 0.2, 0.2} 

\definecolor{sky_blue}{RGB}{37, 166, 213}
\definecolor{light_blue}{rgb}{0.33, 0.80, 1}
\definecolor{dark_blue}{rgb}{0.098, 0.239, 0.52}
\definecolor{ocean}{RGB}{13, 121, 202}
\definecolor{light_ocean}{RGB}{18, 178, 235}
\definecolor{dark_ocean}{RGB}{10, 89, 148}
\definecolor{vibrant_blue}{RGB}{14, 120, 255}

\definecolor{dark_brown}{rgb}{0.3255, 0.004, 0.001}

\newcounter{qnum}
\newcounter{tnum}
\newcommand{\oursc}{\textcolor{org-purp-1}{\textbf{CLIFT}}\xspace}
\newcommand{\ours}{\textcolor{black}{\textbf{CLIFT}}\xspace}

\newcommand{\action}{a}
\newcommand{\acttraj}{\mathbf{\action}}

\DeclareMathOperator*{\argmax}{arg\,max}

\author[1]{Yuxin Chen}
\author[1]{Hari Srikanth}
\author[1]{Nathan Jew}
\author[1]{Menglin Wu}
\author[1]{Pengcheng Wang}
\author[1]{Junli Ren}
\author[1]{Masayoshi Tomizuka}
\author[2]{Peng Xu}
\author[2]{Jinyu Xie}
\author[1,3]{Ran "Thomas" Tian}

\affiliation[1]{University of California, Berkeley}
\affiliation[2]{Google DeepMind}
\affiliation[3]{NVIDIA Research}

\begin{document}

\title{CLIFT: Turning Gemini Robotics On-Device into Humanoid Specialists via
Non-Invasive Closed-Loop Iterative Fine-Tuning}

\abstract{
    While robot foundation models are growing increasingly capable, the strongest models are typically trained on proprietary data and have traditionally remained closed-source, limiting downstream users' ability to adapt them to new tasks, embodiments, and deployment settings.
    Following the LLM community, an emerging access paradigm for closed-weight robot foundation models is the managed supervised fine-tuning (SFT) API, where users submit training data and receive a tuned policy without access to model weights, gradients, or training internals.
    While such APIs let downstream users leverage powerful proprietary foundation models, they restrict policy improvement to pure imitation, ruling out classical reinforcement learning and other closed-loop methods that rely on internal training signals. This limitation is particularly acute for agile, contact-rich humanoid manipulation, where the gap between policy outputs and deployed behavior is large due to novel states, action tracking dynamics, latency, and controller-specific failure modes.
    In this work, we study how effective this new managed-API regime is for humanoid adaptation, and how closed-loop improvement can be realized within it to push policies toward task mastery.
    We conduct one of the first empirical studies of managed-API adaptation on a real humanoid, instantiated on Gemini Robotics On-Device (GROD). We found that direct SFT through the API already substantially outperforms a leading open-weight VLA trained on the same demonstrations, yet still falls short of deployment-level mastery on agile, contact-rich tasks. To close this gap, we introduce \oursc: Closed-Loop Iterative Fine-Tuning, which turns deployment-time reward feedback into API-compatible supervised data and enables closed-loop policy improvement without accessing weights, gradients, likelihoods, or losses---pushing GROD to near-perfect success after two flywheel cycles, all without ``opening the model box.''
}


\website{https://thomaschen98.github.io/clift}{thomaschen98.github.io/clift}

\code{}{coming\_soon}

\maketitle
\begin{figure}[!h]
    \centering
    \includegraphics[width=\linewidth]{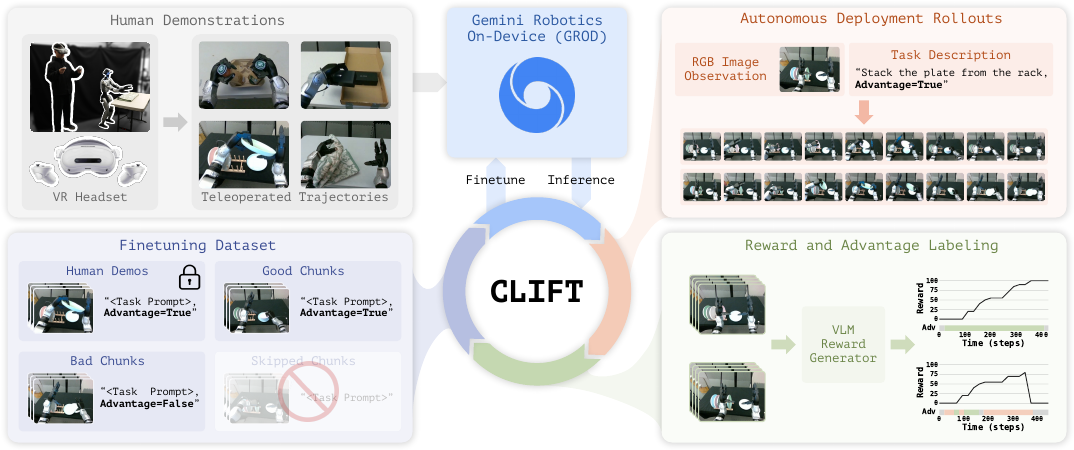}
    \caption{\textbf{\oursc turns \textit{closed-weight} Gemini Robotics into a humanoid specialist.}
    \oursc is an API-only, non-invasive flywheel that bridges managed SFT interfaces and closed-loop policy improvement for humanoid manipulation, without ever touching the model's weights, gradients, or losses.
    Starting from an initial imitation policy, each flywheel cycle deploys the current policy to collect on-device closed-loop \emph{rollouts}, scores them with a preference-calibrated dense reward, and converts each rollout chunk into an SFT tuple carrying a special \emph{advantage token} derived from that reward.
    Fine-tuning on these tuples through GROD's managed SFT API and conditioning on the positive advantage token at deployment yields the next policy, closing the flywheel.}
    \label{fig:teaser}
\end{figure}


\newpage
\section{Introduction}

Foundation vision-language action models (VLAs) trained on large and diverse datasets are producing robot policies with broad competence across tasks and embodiments.
As these foundation VLA models grow stronger, they increasingly determine the ceiling of downstream adaptation: a stronger pretrained VLA can reduce the amount of task-specific data needed and improve generalization beyond the demonstration set~\cite{openx2024, team2025gemini, bjorck2025gr00tn1, wei2026psi0, pi06star, figure_helix2025}.
Yet a fundamental tension underlies this trend: \emph{the strongest pretrained VLAs are also the least accessible}. Many of today's most capable robot foundation models are trained on proprietary robot data and remain as fully closed-source systems~\cite{team2025gemini, figure_helix2025, pi06star}. This creates a practical bind: the models with the highest adaptation ceiling are precisely the ones researchers cannot access.


As the open- versus closed-source debate continues, a middle ground has emerged: \textit{closed-weight} models exposed through managed adaptation interfaces. In the LLM community, SFT APIs already let users customize proprietary models without accessing their weights or training pipelines~\cite{openai2025finetuning,google2025gemini_tuning}.
Robot foundation VLAs are entering this regime, where base-model access matters even more, since robot data is far harder to scale than language data~\cite{tian_icml26}.
Gemini Robotics On-Device (GROD)~\cite{grod2025} is an early example---offering managed fine-tuning while keeping the ``model box'' closed---and Physical Intelligence's partner API~\cite{pi_partner2026} signals the same trend.
This positions closed-weight, API-exposed VLAs as an intermediate regime between fully closed systems and open-weight releases.

These managed-APIs immediately enable fine-tuning on task-specific human demonstrations.
However, for agile, contact-rich \textbf{humanoid tasks}, demonstration-only SFT, which remains the dominant recipe for recent humanoid VLA systems~\cite{humanoid_everyday2025,wei2026psi0}, quickly hits a ceiling.
It trains on a fixed human teleoperation distribution, while deployment occurs under the policy's own closed-loop observation distribution. This mismatch is amplified by two factors. First, execution quality is not captured by task completion alone: smoothness, contact quality, and safety all affect whether a humanoid execution is acceptable. Second, the VLA policy is tightly coupled to the downstream whole-body controller: the same action chunk from a policy can lead to different physical outcomes depending on tracking dynamics, latency, compliance, and contact with the environment~\cite{bronars2026tune}.
Consistent with this limitation, recent humanoid VLA systems trained primarily through demonstration-based SFT remain far from deployment-level success rates on dexterous manipulation suites~\cite{humanoid_everyday2025}.

Crossing this ceiling requires \textbf{on-device closed-loop improvement}: the policy must practice on the actual robot--controller stack, observe its own behavior, and adapt from that feedback rather than from a fixed teleoperation set.
Recent work has pursued this for robot VLAs along several axes---generating residual-RL data for SFT~\cite{xiao2025selfimproving}, running PPO-style RL with learned process rewards~\cite{dopamine}, and steering the policy at test time in its latent action space~\cite{wagenmaker2025steering}.
All of these approaches, however, reach into the learning system: they require auxiliary architectural components (e.g., critic networks or residual policy heads), modifications to the training loss, or access to internal signals such as policy gradients and action likelihoods. None of these levers are exposed by a restricted-access robot foundation model.
This motivates the central question of our work:\vspace{-0.05in}
\begin{center}
    \textit{How effective is the emerging managed-API access regime for humanoid adaptation, and how can closed-loop improvement be realized within this regime to push policies toward task mastery?}
\end{center}

We study this question using Gemini Robotics On-Device~\cite{grod2025} (GROD) on an Unitree G1 humanoid across tasks with varying levels of contact, dexterity, and agility.
We first evaluate direct adaptation by fine-tuning GROD on humanoid demonstrations. The resulting policy substantially outperforms a strong open-weight VLA, $\pi_{0.5}$~\cite{pi05}, trained on the same data, highlighting the value of a stronger closed-weight foundation model even under restricted access. However, a gap to deployment-level mastery remains, particularly on agile, contact-rich manipulation tasks.

To close this gap, we present \oursc: Closed-Loop Iterative Fine-Tuning, a non-invasive closed-loop adaptation pipeline that improves API-only robot foundation models by turning deployment experience into relabeled supervised training data.
Our key insight is that \textbf{reinforcement feedback can be encoded directly into supervised training data}. Specifically, we score deployment rollouts with a preference-calibrated dense reward model and compare action chunks against chunks starting from visually similar states to estimate which behaviors are advantageous. Each rollout chunk is then converted into an SFT tuple: observation, language instruction, action chunk, and a special advantage token that encodes the alignment of the action chunk with the preference. At inference time, conditioning on the positive special advantage steers the policy toward high-quality actions.
In our experiments, two flywheel cycles of CLIFT on top of GROD ultimately push the task success rate close to $100\%$ with emergent behaviors not present in the demonstrations.
As a controlled comparison, we also apply CLIFT to the open-weight $\pi_{0.5}$ model. CLIFT improves this model as well, but its absolute performance remains substantially lower under our protocol, including under the FiLM-style invasive adaptation evaluated in Sec.~\ref{sec:setup}.

\textbf{Contribution statement.} Our work studies an emerging setting at the intersection of two underexplored regimes: \emph{closed-weight robot models}, where a VLA can be adapted only by submitting SFT data to a managed fine-tuning API, and \emph{humanoid closed-loop training}, where VLA actions are tightly coupled to a whole-body controller. This combination amplifies the gap between demonstration learning and deployment behavior. Within this setting, we make the following contributions.
First, we provide one of the first empirical studies of managed-API adaptation for a closed-weight foundation VLA on a real humanoid platform. Direct SFT through the API substantially outperforms an open-weight VLA trained on the same demonstrations, but still falls short of deployment-level mastery on agile, contact-rich tasks.
Second, we introduce \oursc, which turns deployment-time reward feedback into API-compatible supervised data and enables closed-loop policy improvement without accessing weights, gradients, likelihoods, or loss functions, pushing Gemini Robotics On-Device towards task mastery after two flywheel cycles.
This shows that managed SFT APIs can serve not only as customization interfaces, but also as practical interfaces for closed-loop policy improvement---offering a path toward task mastery without ``opening the model box.''
\section{Problem Setting}
\label{sec:formulation}

We access a foundation VLA only through a \emph{managed SFT interface}, a black-box operator $\mathcal{F}_{\mathrm{SFT}}:\; \mathcal{D} \mapsto \pi$ that maps a training dataset $\mathcal{D} = \{(o_t,\, \ell,\, \acttraj_t)\}$ of observation--instruction--action-chunk tuples to a fine-tuned policy $\pi$. It exposes no weights, gradients, losses, or action likelihoods, and is the \emph{only} available update operation. Our loop starts from the demonstration-trained policy $\pi_0 := \mathcal{F}_{\mathrm{SFT}}(\mathcal{D}_{\mathrm{demo}})$, where $\mathcal{D}_{\mathrm{demo}}$ is a fixed set of teleoperated demonstrations.
We aim to improve $\pi_0$ under the closed-loop distribution $p_\pi$ it induces at deployment, using a preference-calibrated dense reward $R_\theta$. Formally, we target the KL-regularized objective
%
\begin{equation}
\pi^* = \argmax_{\pi} \;
\mathbb{E}_{o_{1:T} \sim p_\pi(\cdot \mid \ell)}
\Big[\sum_{t=0}^{T} R_\theta(o_t; \ell)\Big]
- \beta \, \mathbb{E}_{o \sim p_\pi}
\big[\mathrm{KL}\big(\pi(\cdot \mid o,\ell) \,\|\, \pi_0(\cdot \mid o,\ell)\big)\big],
\label{eq:objective}
\end{equation}
%
where $\beta > 0$ anchors $\pi$ to the initialization $\pi_0$ at deployment-visited states.
Standard RL is incompatible with $\mathcal{F}_{\mathrm{SFT}}$: PPO~\cite{schulman2017proximal} needs $\log \pi(a \mid o)$ and backpropagation, and advantage-weighted regression~\cite{peng2019awr} reweights losses by $\exp(A(o,a)/\beta)$---neither is exposed.
We instead approximate $\pi^*$ with a deployment-time data flywheel that operates entirely through $\mathcal{D}$, leaving the model and training procedure untouched.
\section{Closed-Loop Iterative Fine-Tuning}
\label{sec:method}

Our pipeline comprises three steps: \textbf{(i)} bootstrap $\pi_0 = \mathcal{F}_{\mathrm{SFT}}(\mathcal{D}_{\mathrm{demo}})$ from teleoperated demonstrations; \textbf{(ii)} score rollouts with a preference-calibrated dense reward model $R_\theta$ (Sec.~\ref{sec:reward}) and label each chunk with a retrieval-based advantage token (Sec.~\ref{sec:advantage}); \textbf{(iii)} fine-tune the base model on the demonstrations and accumulated relabeled rollouts, then repeat (ii)--(iii) (Sec.~\ref{sec:loop}).

\subsection{Preference-Calibrated Dense Reward Model}
\label{sec:reward}
We seek a dense reward $R_\theta(o_{1:T};\,\ell) \to r_{1:T}$ that scores task progress, execution quality, and safety---the criteria a human operator uses to accept or reject a rollout. 
Existing approaches either prompt a zero-shot VLM~\cite{gvl} or train a reward model from human annotations: the former is scalable but miscalibrated toward surface progress and misses execution failures~\cite{tian_icml26}, while the latter is calibrated but requires prohibitive per-step labeling.
We combine both via a two-stage \emph{select-then-distill} scheme: human preferences select calibrated rewards from VLM-generated candidates, which we then distill into a reusable reward model.
We first build a comparison pool mixing human teleoperated trajectories with initial-SFT-policy rollouts, sample $100$ pairs, and collect a human pairwise preference for each (which rollout better satisfies the task).
For each rollout $o_{1:T}$ in a sampled pair, we prompt a zero-shot VLM~\cite{gvl} with temperature sampling to obtain $K$ candidate per-step reward sequences $\{r_{1:T}^{(k)}\}_{k=1}^{K}$.
We keep the candidates $\{r_{1:T}^*\}$ whose induced cumulative-return rankings best match the human preferences~\cite{tian_icml26, tian2024alignment}, yielding dense per-step rewards that capture execution quality beyond coarse progress.
Finally, we distill $\{r_{1:T}^*\}$ into a generative reward model $R_\theta$~\cite{dopamine} by fine-tuning a VLM on these per-step labels; $R_\theta$ then provides dense, preference-calibrated rewards across all flywheel cycles (more details on the reward model in App.~\ref{app:reward}).

\begin{figure}[t]
    \centering
    \includegraphics[width=\linewidth]{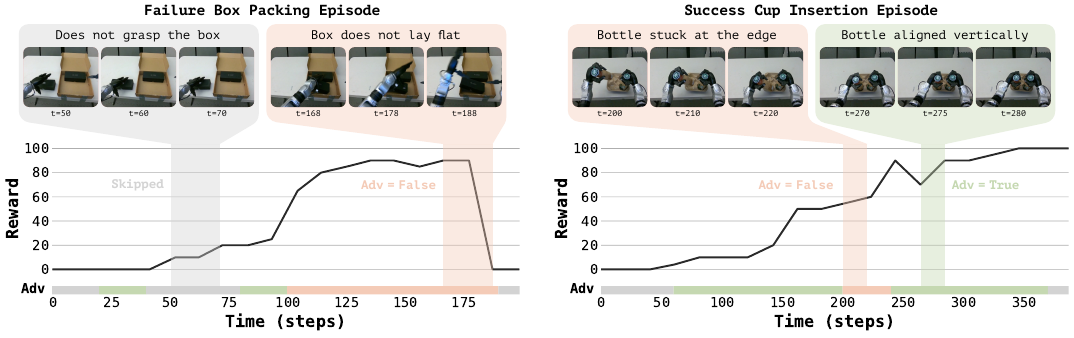}
    \caption{\textbf{Reward predictions and advantage labels on deployment rollouts.} For each rollout, the preference-calibrated reward model $R_\theta$ outputs a sequence of per-step quality scores; these are aggregated into per-chunk returns and binarized into positive/negative advantage labels via retrieval-based comparison against chunks from similar starting states. The labeling operates at the chunk level rather than the episode level: even within a successful rollout, some chunks receive negative labels for sub-optimal execution, while within a failed rollout, well-executed chunks still receive positive labels---providing fine-grained credit assignment that episode-level success/failure annotations cannot capture.}
    \label{fig:reward}
\end{figure}

\subsection{Injecting Reward Signal via Retrieval-Based Advantage Conditioning}
\label{sec:advantage}
As the loop progresses, the policy produces rollouts of varying quality, and the next SFT cycle must know which chunks to amplify and which to suppress. We adopt the \emph{advantage conditioning} paradigm~\cite{pi06star}: each action chunk is augmented with a binary token $I \in \{\text{positive},\, \text{negative}\}$ indicating whether it leads to better outcomes than alternatives from the same state. Both labels are used during SFT, and conditioning on $I = \text{positive}$ at inference steers the policy toward high-quality actions.
Assigning this label requires \emph{state-aware credit assignment}: a chunk must be judged relative to what is achievable from its starting state, since a given return may reflect excellent behavior in a hard recovery state but mediocre behavior in an easy one.

Unlike~\cite{pi06star}, which estimates this advantage by retraining a value model each cycle from a binary success/failure reward---collapsing chunk value to a ``time-to-success'' measure blind to execution quality---we estimate it non-parametrically from our dense reward via retrieval.
For each chunk $\tau_i$ starting at $t_i$ with initial observation $o_{t_i}$, we retrieve, from all rollouts, a comparison set of chunks whose initial observations are visually similar to $o_{t_i}$, and score $\tau_i$ by its discounted return over a look-ahead window of $H$ steps, $G(\tau_i) = \sum_{t=t_i}^{t_i+H} \gamma^{\,t-t_i}\, R_\theta(o_t;\ell)$.
The chunk is labeled \emph{positive} if $G(\tau_i)$ lies in the top $30\%$ of this set and \emph{negative} otherwise; because the comparison is restricted to similar states, the threshold auto-calibrates to state difficulty (details in App.~\ref{app:advantage}).
Fig.~\ref{fig:reward} illustrates this chunk-level labeling on real rollouts.


\subsection{Iterative Self-Improvement}
\label{sec:loop}
At each cycle $k$, we deploy $\pi_k$, score and advantage-label its rollout chunks (Sec.~\ref{sec:advantage}), and append them to a cumulative dataset $\mathcal{D}_k = \mathcal{D}_{\mathrm{demo}} \cup \mathcal{D}_{\mathrm{rollout}}^{1:k}$, with demonstrations always labeled $I = \text{positive}$. Because the managed API fine-tunes the base model from scratch on each submission, we obtain $\pi_{k+1} = \mathcal{F}_{\mathrm{SFT}}(\mathcal{D}_k)$ from the base checkpoint rather than continuing from $\pi_k$---which also avoids compounding distributional drift and re-exploits the full pretrained capacity each cycle. The reward model $R_\theta$ is trained once and held fixed throughout.
\section{Experiments}

We evaluate our method on a real humanoid across a diverse suite of agile, contact-rich tasks. 
Our experiments aim to study three questions: 
(1) Can our closed-loop self-improvement fine-tuning turn a API-only restricted-access foundation model into a humanoid specialist that achieves task mastery?
(2) Does our pipeline generalize across different foundation models with different access regimes?
(3) How does downstream performance compare between GROD and $\pi_{0.5}$ under the same self-improving fine-tuning pipeline?

\begin{figure}[t]
    \centering
    \includegraphics[width=\linewidth]{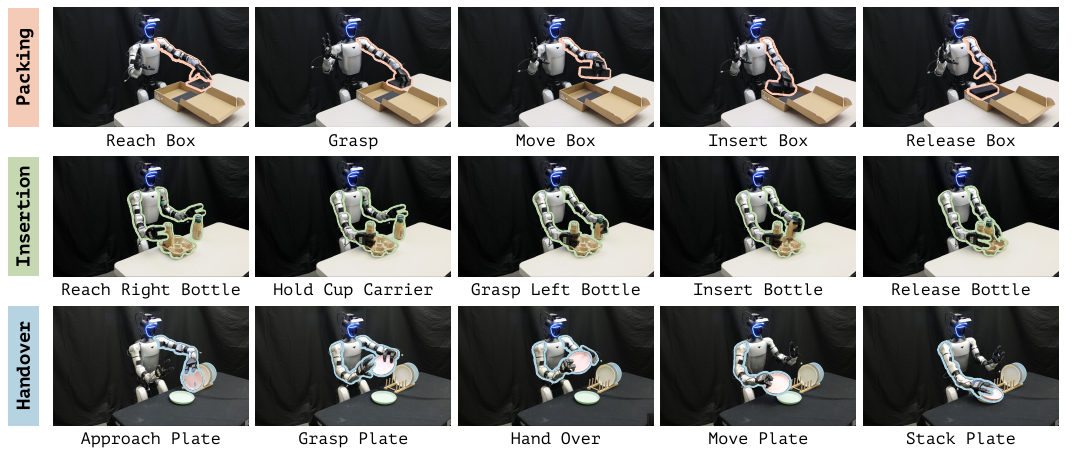}
    \caption{\textbf{Humanoid manipulation task suite.} Three contact-rich Unitree G1 tasks of increasing complexity: \emph{Box Packing} (single-object packing), \emph{Cup Insertion} (precise insertion), and \emph{Bimanual Plate Handover} (coordinated bimanual transfer).}
    \label{fig:tasks}
\end{figure}

\subsection{Experimental Setup}
\label{sec:setup}

\noindent
\textbf{Robot setup.}
We use an Unitree G1 humanoid for experiments. The robot is equipped with dual arms, dexterous hands, and two head-mounted RGB cameras. 
Following~\cite{wei2026psi0}, we adopt a hierarchical whole-body control scheme that decomposes the action space into upper-body and lower-body components: the upper-body action consists of target joint angles for the arms and hands, while the lower-body action specifies planar linear velocities and a target yaw angle. These commands are tracked by an RL-learned whole-body controller~\cite{xr-teleoperate}, which produces the full joint-level commands for execution. At each step, the policy outputs an action chunk spanning $1.6$s into the future and replans after the action chunk is executed. More details can be found in App.~\ref{app:robot_setup}.


\noindent
\textbf{Tasks.}
We evaluate on three contact-rich tasks on the Unitree G1 humanoid (Figure~\ref{fig:tasks}). Although tabletop, they are not fixed-base problems: the humanoid balances with its whole body throughout, so lower-body motions continually shift the camera viewpoint, end-effector pose, and contact geometry seen by the policy.
\textbf{Box Packing}: grasp an object with the three-finger hand and place it in a box, stressing precise grasp formation and release.
\textbf{Cup Insertion}: asymmetric bimanual coordination---one hand stabilizes a cup carrier while the other grasps a bottle and inserts it; small visual, contact, or torso-motion errors make the bottle collide, tilt, or miss.
\textbf{Bimanual Plate Handover}: pick up a plate with one hand, transfer it to the other, and place it at a target, demanding precise inter-arm timing and stable finger contacts despite torso and base motion.

\begin{wrapfigure}{r}{0.45\linewidth}
\centering
\includegraphics[width=\linewidth]{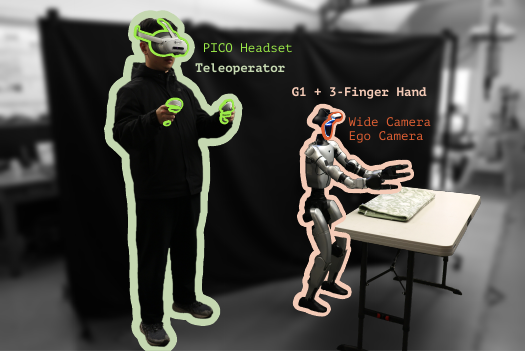}
\caption{\textbf{Whole-body teleoperation setup.} The operator wears a VR headset, two handheld controllers, and two ankle IMU trackers, which together yield a real-time full-body SMPL pose that the whole-body controller~\cite{xr-teleoperate} maps to robot joint commands.}
\label{fig:teleop}
\vspace{-1em}
\end{wrapfigure}

\noindent
\textbf{Pre-trained VLA models for humanoid adaptation.}
We apply \oursc to two pre-trained VLA models that represent different access regimes. Gemini Robotics On-Device~\cite{grod2025} is a closed-weight model exposed through a managed SFT API, 
while \textbf{$\pi_{0.5}$}~\cite{pi05} is an open-weight VLA model released by Physical Intelligence, allowing direct access to model weights and training code.

To enable a controlled comparison, we use the same teleoperated demonstrations, the same rollout collection budget at each flywheel cycle, the same preference-calibrated reward model $R_\theta$, and the reward injection procedure for both models.
Both models are fine-tuned with a standard supervised loss on observation-action pairs for 10 epochs per flywheel cycle using their default fine-tuning configurations, without model-specific hyperparameter tuning. For each model, we perform two independent training runs and select the run with the lowest held-out demonstration loss as the final model. While the two foundation models also differ in architecture and scale, which our experiments cannot independently isolate, these controls ensure that any observed performance gap is not driven by mismatches in our adaptation training data, optimization procedure, or evaluation protocol.

\noindent
\textbf{Human demonstration collection.}
We collect demonstrations via whole-body VR teleoperation (Figure~\ref{fig:teleop}). 
During each demonstration, we record the egocentric camera image, proprioceptive state, and executed actions to form training trajectories. For each task, we collect 2 hours of human teleoperated data. More details on demonstration collection can be found in App.~\ref{app:robot_setup}.

\noindent
\textbf{\oursc protocol and evaluation.}
We initialize $\pi_0$ by fine-tuning the base model on 2 hours of teleoperated demonstrations per task, and train the preference-calibrated reward model from $100$ human pairwise comparisons over these demonstrations and initial SFT rollouts (Sec.~\ref{sec:reward}). We then run $2$ cycles of \oursc (Sec.~\ref{sec:loop}). A key property of the on-device setting is that deployment and evaluation coincide: each cycle deploys the current policy for $100$ rollouts per task under a fixed evaluation suite, and these same rollouts both define its success rate and, once relabeled, form the training data for the next cycle. More details on evaluation can be found in App.~\ref{app:evaluation}.

\subsection{Results}

\begin{figure}[t]
    \centering
    \includegraphics[width=\linewidth]{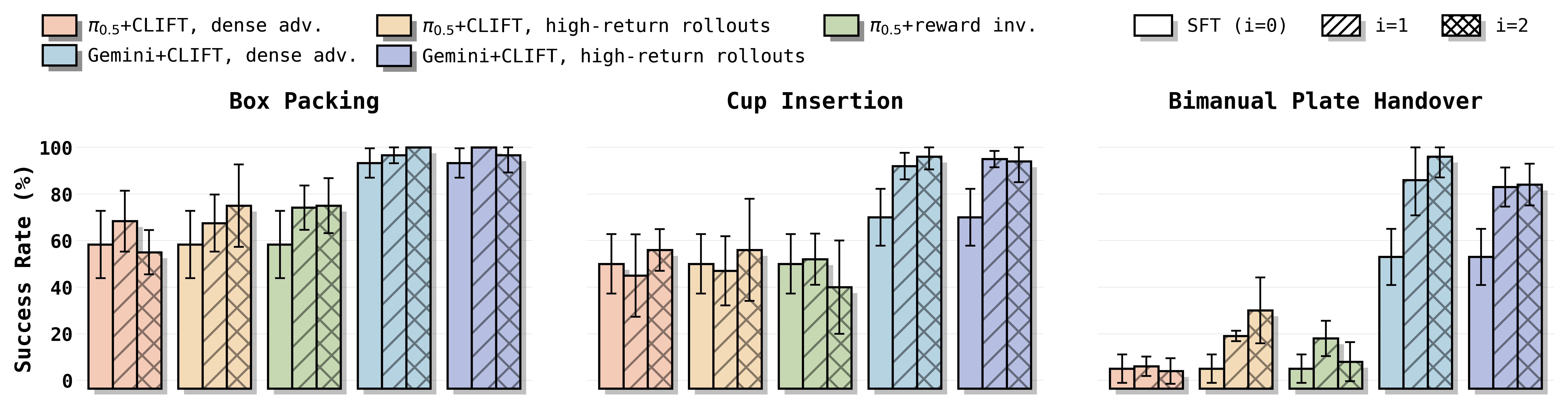}
    \caption{\textbf{\oursc lifts humanoid manipulation policies across foundation models and access regimes.} Success rate (\%, over $100$ trials) across two flywheel cycles on three humanoid tasks. Applied to Gemini Robotics On-Device, both \oursc iteratively improves the demonstration-trained baseline, reaching higher performance at each cycle. The same pipeline also improves the open-weight $\pi_{0.5}$, but plateaus at a substantially lower ceiling. An \emph{invasive} adaptation of $\pi_{0.5}$ that injects the preference signal directly via FiLM-style conditioning (an architectural change) still does not close the gap to GROD.}
    \label{fig:results}
\end{figure}

\noindent
\textbf{CLIFT effectively turns GROD into a humanoid specialist that achieves task mastery.}
We apply \oursc to GROD following the protocol described in Sec.~\ref{sec:setup}, and evaluate two API-compatible variants.
In the \emph{dense advantage-conditioned} variant, every action chunk in the deployment rollouts is scored by the preference-calibrated reward model, assigned a positive or negative advantage label via retrieval-based comparison, and included in the next SFT dataset.
In the \emph{high-return episode selection} variant, we use a DAgger-style data selection rule: deployment rollouts are ranked by accumulated reward, and only advantageous rollouts (top $30\%$) are added to the next SFT dataset.
The two variants differ along two axes: the dense variant operates at the \emph{chunk} level and conditions on positive and negative advantage labels, whereas the episode-selection variant operates at the \emph{rollout} level, performing filtered imitation on the top-$30\%$ highest-return rollouts and discarding the rest.

Figure~\ref{fig:results} reports GROD's success rate over the two flywheel cycles. Both variants improve the demonstration-trained policy at every cycle on all three tasks. The dense advantage-conditioned variant is strongest, lifting GROD from its SFT baseline to near-task mastery: $93\%\!\to\!100\%$ on Box Packing, $70\%\!\to\!98\%$ on Cup Insertion, and $53\%\!\to\!96\%$ on Bimanual Plate Handover---all through API-only access, with no additional human demonstrations.

Interestingly, the two variants are comparable on Box Packing and Cup Insertion, both approaching perfect success, but diverge sharply on the hardest task, Bimanual Plate Handover, where the dense variant reaches $96\%$ versus only $\sim\!84\%$ for episode selection.
The two variants collect the same rollouts and differ only in what enters the next SFT dataset: episode selection keeps the top-$30\%$ highest-return rollouts and \emph{discards the rest}, whereas the dense variant retains every rollout and labels its chunks by advantage.
On easy tasks, successful rollouts are plentiful, so discarding the low-return ones costs little and the two variants match. On hard tasks, however, successes are rare and most collected rollouts are failures---precisely the data that episode selection throws away.
By instead retaining these failed rollouts and relabeling them at the chunk level, the dense variant recovers well-executed chunks as positive examples and exploits poorly-executed ones as contrastive negatives, converting failure into usable training signal (Fig.~\ref{fig:reward}).
The gain on Bimanual Plate Handover therefore comes specifically from learning from failed rollouts, rather than from successful trajectories alone.


\noindent
\textbf{CLIFT can adapt different foundation models to humanoid tasks.}
To test whether \oursc generalizes beyond GROD, we apply the identical pipeline---the same reward model $R_\theta$, rollout budget, and advantage relabeling---to the open-weight $\pi_{0.5}$~\cite{pi05}, which differs from GROD in architecture, pretraining scale, and access regime.
\oursc improves $\pi_{0.5}$ as well: over two flywheel cycles, success rate rises from $59\%$ to $76\%$ on Box Packing, $50\%$ to $56\%$ on Cup Insertion, and $5\%$ to $30\%$ on Bimanual Plate Handover.
The same non-invasive mechanism therefore transfers across foundation models and access regimes, even though the absolute performance $\pi_{0.5}$ attains remains well below GROD---a gap we examine next.

\begin{figure}[t]
    \centering
    \includegraphics[width=\linewidth]{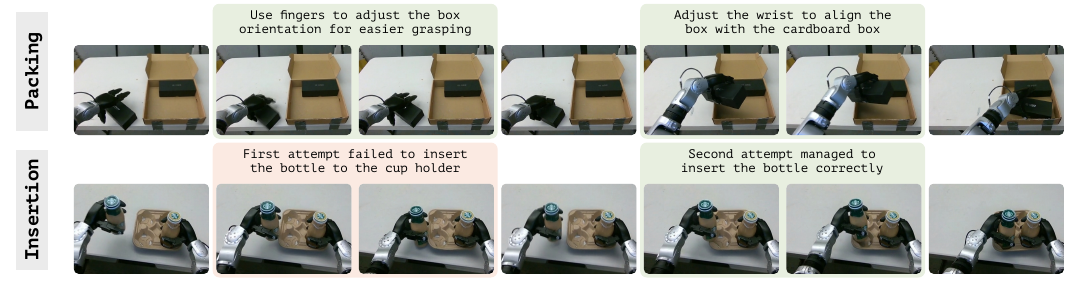}
    \caption{\textbf{Emergent behaviors after closed-loop fine-tuning.} By the final flywheel cycle, GROD exhibits behaviors not present in the teleoperated demonstrations. \emph{Top (Box Packing):} the policy reorients the box with its fingers and wrist to set up an easier grasp before lifting. \emph{Bottom (Cup Insertion):} the policy recovers from a failed first insertion attempt and succeeds on a second try, rather than aborting.}
    \label{fig:qualitative}
\end{figure}

\noindent
\textbf{Foundation model choice matters, even under API-only adaptation.}
\oursc improves both foundation models, but the absolute ceiling it reaches is set by the strength of the pretrained prior. $\pi_{0.5}$ builds on a PaliGemma backbone~\cite{paligemma}, whereas GROD builds on Gemma, pretrained on substantially larger and more diverse corpora. With the same pipeline and the same demonstrations, the policy initialized from the more extensively pretrained model attains higher success at \emph{every} flywheel cycle, and the gap widens with task difficulty: on the most agile, contact-rich tasks, Gemini reaches $84\%$ versus $\pi_{0.5}$'s $46\%$ on Bimanual Plate Handover, and $88\%$ versus $46\%$ on Cup Insertion.
Critically, this gap does not stem from the API-only restriction. Even when $\pi_{0.5}$ is granted \emph{invasive} adaptation that injects the preference signal directly into the network via FiLM-style conditioning~\cite{film} (more details in App.~\ref{app:additional_result})---a lever \oursc deliberately forgoes---it still trails Gemini adapted through the API alone (e.g., $40\%$ vs.\ $84\%$ on Bimanual Plate Handover, $48\%$ vs.\ $93\%$ on Cup Insertion). A stronger model adapted through a narrow SFT interface thus outperforms a weaker model granted full internal access.
We evaluate a single invasive baseline and do not claim to rule out the broader space of full-access methods. Still, the pattern is consistent: the gap tracks pretrained capability rather than the level of training access. This is precisely the positioning of our introduction---when closed-loop adaptation is the bottleneck, the choice of base model can matter as much as the access regime, which is why securing API access to the strongest closed-weight foundation models is worthwhile even when their internals remain hidden.
Though we only showed three representative tasks in Fig.~\ref{fig:qualitative}, a detailed benchmarking between GROD and $\pi_{0.5}$ can be found in this interactive server at our \href{https://thomaschen98.github.io/clift}{\textcolor{orange}{Project Website}}.

\noindent
\textbf{Closed-loop practice elicits emergent behaviors absent from demonstrations.}
The advantage of a stronger base model also shows up qualitatively. By the final flywheel cycle, GROD acquires corrective and pre-manipulation behaviors that never appear in the teleoperated demonstrations (Fig.~\ref{fig:qualitative}). On Box Packing, the policy learns to first nudge and reorient the box with its fingers and wrist to set up an easier grasp, rather than grasping it in whatever pose it starts in. On Cup Insertion, it acquires a retry behavior: when the first insertion attempt fails, the policy re-approaches and succeeds on a second attempt instead of stalling. Because \oursc reinforces high-return chunks recovered from the policy's \emph{own} rollouts---including chunks salvaged from failed episodes---closed-loop practice does more than imitate demonstrated motions; it composes new behavior beyond the demonstration distribution, and a stronger pretrained prior is better able to acquire it.

\section{Conclusion}

While robot foundation models grow increasingly capable, the strongest are typically trained on proprietary data and exposed only through managed supervised fine-tuning (SFT) APIs---users submit training data and receive a tuned policy, with no access to model weights, gradients, or training internals. In this work, we studied how effective this managed-API regime is for humanoid adaptation, and how closed-loop improvement can be realized within it to push policies toward task mastery.
We provided one of the first empirical studies of managed-API adaptation on a real humanoid, instantiated on Gemini Robotics model. Direct SFT through the API already substantially outperforms a leading open-weight VLA trained on the same demonstrations, yet still falls short of deployment-level mastery on agile, contact-rich tasks. To close this gap, we introduced \oursc: Closed-Loop Iterative Fine-Tuning, which turns deployment-time reward feedback into API-compatible supervised data and enables closed-loop policy improvement without accessing weights, gradients, likelihoods, or losses-pushing Gemini Robotics model to near-perfect success after two flywheel cycles, all without ``opening the model box.''

\textbf{Limitations.} \ours relies on real-robot rollouts at each cycle to inject real-world deployment signal into the policy. This signal is especially valuable for complex embodiments, where coupling to a whole-body controller induces a large gap between a policy's predicted actions and its deployed behavior; at the same time, such rollouts are costly and safety-sensitive to collect on hardware. We are excited to integrate a control-aware world model that anticipates deployment outcomes, reducing the number of risky real-robot rollouts required per cycle. In addition, we only evaluated one open-weight model and will expand the comparison to more open-weight models in future work.

\clearpage
\bibliographystyle{unsrt}
\bibliography{references.bib}

@inproceedings{film,
  title     = {{FiLM}: Visual Reasoning with a General Conditioning Layer},
  author    = {Perez, Ethan and Strub, Florian and de Vries, Harm and Dumoulin, Vincent and Courville, Aaron},
  booktitle = {AAAI Conference on Artificial Intelligence},
  year      = {2018}
}

@article{humanoid_everyday2025,
  title         = {Humanoid Everyday: A Comprehensive Robotic Dataset for Open-World Humanoid Manipulation},
  author        = {Zhao, Zhenyu and Jing, Hongyi and Liu, Xiawei and Mao, Jiageng and Jha, Abha and Yang, Hanwen and Xue, Rong and Zakharov, Sergey and Guizilini, Vitor and Wang, Yue},
  journal       = {arXiv preprint arXiv:2510.08807},
  year          = {2025},
  archivePrefix = {arXiv},
  eprint        = {2510.08807}
}

@inproceedings{gvl,
  title     = {Vision Language Models are In-Context Value Learners},
  author    = {Ma, Yecheng Jason and Hejna, Joey and Fu, Chuyuan and Shah, Dhruv and Liang, Jacky and Xu, Zhuo and Kirmani, Sean and Xu, Peng and Driess, Danny and Xiao, Ted and Bastani, Osbert and Jayaraman, Dinesh and Yu, Wenhao and Zhang, Tingnan and Sadigh, Dorsa and Xia, Fei},
  booktitle = {International Conference on Learning Representations (ICLR)},
  year      = {2025}
}

@inproceedings{tian_icml26,
  title     = {Position: Good Embodied Reward Models Need Bad Behavior Data},
  author    = {Tian, Ran and Wu, Yilin and Bajcsy, Andrea},
  booktitle = {International Conference on Machine Learning (ICML)},
  year      = {2026},
  url       = {https://cmu-intentlab.github.io/pdf/tian_icml_26_position.pdf}
}

@article{tian2024alignment,
  title         = {Maximizing Alignment with Minimal Feedback: Efficiently Learning Rewards for Visuomotor Robot Policy Alignment},
  author        = {Tian, Ran and Wu, Yilin and Xu, Chenfeng and Tomizuka, Masayoshi and Malik, Jitendra and Bajcsy, Andrea},
  journal       = {arXiv preprint arXiv:2412.04835},
  year          = {2024},
  archivePrefix = {arXiv},
  eprint        = {2412.04835}
}

@article{bronars2026tune,
  title         = {Tune to Learn: How Controller Gains Shape Robot Policy Learning},
  author        = {Bronars, Antonia and Park, Younghyo and Agrawal, Pulkit},
  journal       = {arXiv preprint arXiv:2604.02523},
  year          = {2026},
  archivePrefix = {arXiv},
  eprint        = {2604.02523}
}

@article{xiao2025selfimproving,
  title         = {Self-Improving Vision-Language-Action Models with Data Generation via Residual {RL}},
  author        = {Xiao, Wenli and Lin, Haotian and Peng, Andy and Xue, Haoru and He, Tairan and Xie, Yuqi and Hu, Fengyuan and Wu, Jimmy and Luo, Zhengyi and Fan, Linxi and Shi, Guanya and Zhu, Yuke},
  journal       = {arXiv preprint arXiv:2511.00091},
  year          = {2025},
  archivePrefix = {arXiv},
  eprint        = {2511.00091}
}

@article{dopamine,
  title         = {{Robo-Dopamine}: General Process Reward Modeling for High-Precision Robotic Manipulation},
  author        = {Tan, Huajie and others},
  journal       = {arXiv preprint arXiv:2512.23703},
  year          = {2025},
  archivePrefix = {arXiv},
  eprint        = {2512.23703}
}

@article{cfgrl,
  title         = {Diffusion Guidance Is a Controllable Policy Improvement Operator},
  author        = {Frans, Kevin and Park, Seohong and Abbeel, Pieter and Levine, Sergey},
  journal       = {arXiv preprint arXiv:2505.23458},
  year          = {2025},
  archivePrefix = {arXiv},
  eprint        = {2505.23458}
}

@article{paligemma,
  title         = {{PaliGemma}: A Versatile 3{B} {VLM} for Transfer},
  author        = {Beyer, Lucas and Steiner, Andreas and Pinto, Andr{\'e} Susano and Kolesnikov, Alexander and Wang, Xiao and Salz, Daniel and Neumann, Maxim and Alabdulmohsin, Ibrahim and Tschannen, Michael and Bugliarello, Emanuele and others},
  journal       = {arXiv preprint arXiv:2407.07726},
  year          = {2024},
  archivePrefix = {arXiv},
  eprint        = {2407.07726}
}

@misc{xr-teleoperate,
  author       = {{Unitree Robotics}},
  title        = {{XR-Teleoperate}: An Open-Source Teleoperation Framework and Data Collection Toolkit for Embodied Intelligence},
  howpublished = {\url{https://github.com/unitreerobotics/xr_teleoperate}},
  year         = {2024},
  note         = {Accessed: 2026-02}
}

@misc{openai2025finetuning,
  author = {OpenAI},
  title = {Fine-tuning -- OpenAI API Documentation},
  year = {2025},
  url = {https://platform.openai.com/docs/guides/fine-tuning/},
  note = {Accessed: 2026-04-13}
}

@article{schulman2017proximal,
  title={Proximal policy optimization algorithms},
  author={Schulman, John and Wolski, Filip and Dhariwal, Prafulla and Radford, Alec and Klimov, Oleg},
  journal={arXiv preprint arXiv:1707.06347},
  year={2017}
}

@article{peng2019awr,
  title={Advantage-Weighted Regression: Simple and Scalable Off-Policy Reinforcement Learning},
  author={Peng, Xue Bin and Kumar, Aviral and Zhang, Grace and Levine, Sergey},
  journal={arXiv preprint arXiv:1910.00177},
  year={2019},
  archivePrefix={arXiv},
  eprint={1910.00177}
}

@article{wagenmaker2025steering,
  title={Steering your diffusion policy with latent space reinforcement learning},
  author={Wagenmaker, Andrew and Nakamoto, Mitsuhiko and Zhang, Yunchu and Park, Seohong and Yagoub, Waleed and Nagabandi, Anusha and Gupta, Abhishek and Levine, Sergey},
  journal={arXiv preprint arXiv:2506.15799},
  year={2025}
}

@misc{pi_partner2026,
  author = {Physical Intelligence},
  title = {The Physical Intelligence Layer},
  year = {2026},
  month = feb,
  url = {https://www.pi.website/blog/partner},
  note = {Accessed: 2026-04-13}
}

@misc{grod2025,
  author = {Google DeepMind},
  title = {Gemini Robotics On-Device},
  year = {2025},
  url = {https://deepmind.google/models/gemini-robotics/gemini-robotics-on-device/},
  note = {Private preview, available via Trusted Tester Program. Accessed: 2026-04-13}
}

@misc{google2025gemini_tuning,
  author = {Google Cloud},
  title = {About supervised fine-tuning for {Gemini} models -- {Vertex AI} Documentation},
  year = {2025},
  url = {https://docs.cloud.google.com/vertex-ai/generative-ai/docs/models/gemini-supervised-tuning},
  note = {Accessed: 2026-04-13}
}

@article{team2025gemini,
  title={Gemini robotics: Bringing ai into the physical world},
  author={Team, Gemini Robotics and Abeyruwan, Saminda and Ainslie, Joshua and Alayrac, Jean-Baptiste and Arenas, Montserrat Gonzalez and Armstrong, Travis and Balakrishna, Ashwin and Baruch, Robert and Bauza, Maria and Blokzijl, Michiel and others},
  journal={arXiv preprint arXiv:2503.20020},
  year={2025}
}

@article{wei2026psi0,
  title   = {$\Psi_0$: An Open Foundation Model Towards Universal Humanoid Loco-Manipulation},
  author  = {Wei, Songlin and Jing, Hongyi and Li, Boqian and Zhao, Zhenyu and Mao, Jiageng and Ni, Zhenhao and He, Sicheng and Liu, Jie and Liu, Xiawei and Kang, Kaidi and others},
  journal = {arXiv preprint arXiv:2603.12263},
  year    = {2026}
}

@inproceedings{openx2024,
  title     = {Open {X}-Embodiment: Robotic Learning Datasets and {RT-X} Models},
  author    = {{Open X-Embodiment Collaboration}},
  booktitle = {IEEE International Conference on Robotics and Automation (ICRA)},
  pages     = {6892--6903},
  year      = {2024},
  doi       = {10.1109/ICRA57147.2024.10611477}
}

@misc{figure_helix2025,
  title  = {Helix: A Vision-Language-Action Model for Generalist Humanoid Control},
  author = {{Figure AI}},
  year   = {2025},
  url    = {https://www.figure.ai/helix}
}

@inproceedings{pi05,
  title     = {$\pi_{0.5}$: a Vision-Language-Action Model with Open-World Generalization},
  author    = {Black, Kevin and Brown, Noah and Darpinian, James and Dhabalia, Karan and Driess, Danny and Esmail, Adnan and Equi, Michael and Finn, Chelsea and Fusai, Niccolo and Galliker, Manuel Y. and Ghosh, Dibya and Groom, Lachy and Hausman, Karol and Ichter, Brian and Jakubczak, Szymon and Jones, Tim and Ke, Liyiming and LeBlanc, Devin and Levine, Sergey and others},
  booktitle = {Proceedings of the 9th Conference on Robot Learning (CoRL)},
  pages     = {17--40},
  volume    = {305},
  publisher = {PMLR},
  year      = {2025}
}

@article{pi06star,
  title         = {$\pi^*_{0.6}$: a {VLA} That Learns From Experience},
  author        = {Amin, Ali and Aniceto, Raichelle and Balakrishna, Ashwin and Black, Kevin and Conley, Ken and Connors, Gavin and Darpinian, James and Dhabalia, Karan and DiCarlo, Joseph and Driess, Danny and others},
  journal       = {arXiv preprint arXiv:2511.14759},
  year          = {2025},
  archivePrefix = {arXiv},
  eprint        = {2511.14759}
}

@article{bjorck2025gr00tn1,
  title         = {GR00T N1: An Open Foundation Model for Generalist Humanoid Robots},
  author        = {Bjorck, Johan and Casta{\~n}eda, Fernando and Cherniadev, Nikita and Da, Xingye and Ding, Runyu and Fan, Linxi and Fang, Yu and Fox, Dieter and Hu, Fengyuan and Huang, Spencer and Jang, Joel and Jiang, Zhenyu and Kautz, Jan and Kundalia, Kaushil and Lao, Lawrence and Li, Zhiqi and Lin, Zongyu and Lin, Kevin and Liu, Guilin and Llontop, Edith and Magne, Loic and Mandlekar, Ajay and Narayan, Avnish and Nasiriany, Soroush and Reed, Scott and Tan, You Liang and Wang, Guanzhi and Wang, Zu and Wang, Jing and Wang, Qi and Xiang, Jiannan and Xie, Yuqi and Xu, Yinzhen and Xu, Zhenjia and Ye, Seonghyeon and Yu, Zhiding and Zhang, Ao and Zhang, Hao and Zhao, Yizhou and Zheng, Ruijie and Zhu, Yuke},
  journal       = {arXiv preprint arXiv:2503.14734},
  year          = {2025},
  archivePrefix = {arXiv},
  eprint        = {2503.14734}
}

@article{simeoni2025dinov3,
  title={Dinov3},
  author={Sim{\'e}oni, Oriane and Vo, Huy V and Seitzer, Maximilian and Baldassarre, Federico and Oquab, Maxime and Jose, Cijo and Khalidov, Vasil and Szafraniec, Marc and Yi, Seungeun and Ramamonjisoa, Micha{\"e}l and others},
  journal={arXiv preprint arXiv:2508.10104},
  year={2025}
}

\clearpage
\beginappendix{
\section{Related Work}
\label{app: related_work}
\noindent
\textbf{Closed-weight API-exposed foundation models}.
Despite the rapid progress of open-source models, many of the most capable large language models and vision-language models remain closed-weight, as they are typically trained on proprietary datasets and supported by large-scale proprietary training infrastructure. As a result, downstream users often lack direct access to model weights and training pipelines.
While the debate between open-source and closed-source models continues, the community has increasingly explored ways for downstream users to benefit from powerful closed-source models beyond inference alone. One emerging path is to expose managed adaptation interfaces, such as supervised fine-tuning APIs, that enable limited customization without releasing model weights or full training pipelines.
Robot foundation models are beginning to enter this regime as well, but in a form where access to strong base models is arguably even more critical. Gemini Robotics On-Device~\cite{grod2025} and Physical Intelligence's partner API~\cite{pi_partner2026} are early examples of this model form in robotics: rather than exposing model weights, they provide managed fine-tuning interfaces that allow downstream users to adapt the underlying model on custom data while keeping the model internals and training pipeline inaccessible. This positions both as an intermediate model-access regime in robotics, bridging fully closed proprietary systems and open-weight robot foundation models.
In this paper, we conduct the first study of how a closed-weight robot foundation model, accessed only through a managed fine-tuning API, can be specialized for challenging humanoid deployment settings.

\noindent
\textbf{Vision-Language-Action models for humanoid manipulation}.
Vision-language-action (VLA) models have emerged as a dominant paradigm for instruction-following robot policies, leveraging large-scale pre-trained vision-language backbones to enable broad task and embodiment generalization~\cite{pi05, openx2024}.
More recently, this paradigm has been extended to humanoid platforms, with models such as GR00T N1~\cite{bjorck2025gr00tn1}, $\Psi_0$~\cite{wei2026psi0}, and Helix~\cite{figure_helix2025} bringing VLA-style architectures to whole-body control, supported by growing efforts toward large-scale humanoid manipulation datasets~\cite{humanoid_everyday2025}.
However, these models are typically trained through supervised behavior cloning on expert teleoperation data---an imitation-only paradigm that fundamentally caps performance at demonstration quality and coverage. For agile, contact-rich humanoid tasks, where expert demonstrations are expensive to collect at scale and inherently noisy, this leaves a substantial gap between training and deployment-level mastery.
In the paper, we study how to push humanoid VLAs beyond this demonstration-trained ceiling through deployment-time self-improvement, particularly under the constraint of API-only adaptation.

\noindent
\textbf{On-device closed-loop policy improvement for robot VLAs}.
Imitation learning alone is often insufficient for robot deployment because it trains on a fixed demonstration distribution while inference occurs under the policy's own closed-loop state distribution. This mismatch becomes more severe when preferences extend beyond task completion to execution quality, safety, and interaction with the low-level controller: the same high-level action can lead to different physical outcomes depending on tracking dynamics, latency, compliance, and contact with the environment~\cite{bronars2026tune}. These issues are particularly salient for humanoids, where a VLA policy must coordinate dexterous hands and upper-body motion while acting through a downstream whole-body controller.
Closed-loop policy improvement addresses this mismatch by letting the policy practice, observe and learn from its own failures and partial successes. Recent work has shown that deployment experience can improve robot VLAs through residual-RL data generation~\cite{xiao2025selfimproving}, PPO-style RL with learned process rewards~\cite{dopamine}, and latent-space test-time policy steering~\cite{wagenmaker2025steering}. 
However, these approaches typically assume invasive access to the learning system, such as trainable critics or residual policies, loss reweighting, policy gradients, or action likelihoods. They also primarily target embodiments whose action interfaces are substantially simpler: fixed-base manipulators or low-DoF grippers where the gap between the VLA action chunk prediction and the deployment-time result is comparatively narrower.
In contrast, our work studies closed-loop policy improvement at the intersection of two emerging but underexplored regimes: a restricted model-access regime, where the only available update operation is managed SFT, and a complex embodiment regime, where humanoid policies must coordinate dexterous manipulation through a whole-body controller. We express the improvement signal entirely as SFT-compatible data, enabling deployment-time self-improvement without modifying model internals.

\section{Reward Model Implementation Details}
\label{app:reward}

We seek a dense reward $R_\theta(o_{1:T};\,\ell) \to r_{1:T}$ that scores task progress, execution quality, and safety---the criteria a human operator uses to accept or reject a rollout. 
Existing approaches either prompt a zero-shot VLM~\cite{gvl} or train a reward model from human annotations: the former is scalable but miscalibrated toward surface progress and misses execution failures~\cite{tian_icml26}, while the latter is calibrated but requires prohibitive per-step labeling.
We combine both via a two-stage \emph{select-then-distill} scheme: human preferences select calibrated rewards from VLM-generated candidates, which we then distill into a reusable reward model.

\paragraph{Collecting human preferences.}
To calibrate the reward model, we collect a small set of human pairwise preferences. We build a comparison pool that mixes human teleoperated trajectories (including failed attempts) with rollouts from the initial SFT policy, so that it spans the full spectrum of behavior quality, and sample $100$ rollout pairs from it. Each pair is presented to a single expert evaluator as a synchronized, side-by-side replay from three viewpoints---the robot's two egocentric head cameras (regular and wide-angle) and an external third-person view---so that both the fine manipulation and the whole-body motion are visible. Given the task description, the evaluator marks the rollout that better accomplishes the task while also exhibiting cleaner grasps, smoother motion, and safer contacts, yielding $100$ rollout-level pairwise preference labels.

\paragraph{Generating and selecting preference-consistent rewards.}
To obtain dense per-step rewards for the human-labeled rollouts, we use a VLM as a generator of candidate reward sequences, following the generative reward-modeling framework of~\cite{gvl}. For each rollout, we prompt the VLM (GPT-5.5) with the task description and the rollout frames and draw $K = 12$ candidate per-step reward sequences $\{r_{1:T}^{(k)}\}_{k=1}^{12}$ under temperature sampling. The full prompt template is shown below, with task-specific content filled into the placeholders \texttt{\{TASK\_INSTRUCTION\}}, \texttt{\{RUBRIC\}}, and \texttt{\{HARD\_FAILURES\}}.

\begin{tcolorbox}[breakable, colback=gray!5, colframe=gray!50, title= Generative Reward Modeling Template]
\small\ttfamily
You are an expert reward labeler for robot manipulation videos.\\[2pt]

TASK: \{TASK\_INSTRUCTION\}\\[2pt]

REWARD RUBRIC (anchor scores in [0, 100]; interpolate between them):\\
- ~~0: No relevant action; robot has not engaged with the task\\
\{RUBRIC\}\\[2pt]

HARD FAILURES (override the rubric --- if any condition below is visible in a frame, that frame's task\_reward MUST be 0, no matter what milestone was reached earlier):\\
\{HARD\_FAILURES\}\\[2pt]

You will be shown $N$ frames sampled from a robot rollout. The order has been shuffled --- ignore frame order and judge each frame purely on its visual content. Each frame is labeled with a ``frame\_index'' matching the order shown.\\[2pt]

Now label these $N$ frames of the rollout and the reason for the score:\\
Frame 1: \textlangle image\textrangle\\
Frame 2: \textlangle image\textrangle\\
\dots\\
Frame $N$: \textlangle image\textrangle\\[2pt]

OUTPUT FORMAT --- return a single JSON array of length $N$ with one object per frame, in the SAME ORDER the frames were shown:\\
~~[\\
~~~~\{"frame\_index": 1, "task\_reward": \textlangle int 0..100\textrangle, "frame\_description": "\textlangle one short sentence\textrangle"\},\\
~~~~\{"frame\_index": 2, "task\_reward": \textlangle int 0..100\textrangle, "frame\_description": "\textlangle one short sentence\textrangle"\},\\
~~~~\dots\\
~~]\\
Output ONLY the JSON array. No prose, no markdown fences.
\end{tcolorbox}

\paragraph{Placeholder semantics.} Each placeholder captures one axis of user-specified, task-level preference:
\begin{itemize}
  \item \texttt{\{TASK\_INSTRUCTION\}}: a one- or two-sentence natural-language description of the task goal.
  \item \texttt{\{RUBRIC\}}: a list of progress milestones, each tied to an anchor score in $[0, 100]$. The model linearly interpolates between anchors for intermediate frames.
  \item \texttt{\{HARD\_FAILURES\}}: a list of visually-checkable failure conditions that override the rubric and force $r_t = 0$, used to prevent partial credit for frames where a milestone was momentarily reached but the resulting state is actually a failure.
\end{itemize}

Each of the $K = 12$ candidates assigns its own rollout a cumulative return; a single candidate therefore scores only the rollout it was generated for, not the entire pair set. Rather than labeling each pair independently, we choose one candidate sequence \emph{per rollout} so as to maximize the number of human-labeled pairs whose induced returns agree with the preference---writing $G_X^{(k)}$ for the return of candidate $k$ on rollout $X$, a pair $A \succ B$ agrees when the selected $G_A > G_B$~\cite{tian_icml26, tian2024alignment}. The resulting dense per-step labels $\{r_{1:T}^{*}\}$ are the distillation targets for the reward model $R_\theta$ trained next.

\paragraph{Reward model training.}
We instantiate $R_\theta$ as a vision--language model that takes the rollout frames together with the task instruction and predicts a per-step scalar reward. We use Qwen3-VL as the backbone and regress its predicted per-step rewards onto the preference-selected dense labels $\{r_{1:T}^{*}\}$ under a mean-squared error loss. We pool the labeled data from \emph{all} tasks into a single training set and train one shared reward model rather than a per-task model: this lets $R_\theta$ exploit notions of grasp quality, smoothness, and safety that transfer across tasks, and amortizes the human-preference budget over the whole suite. The same $R_\theta$ then scores rollouts for every task and is held fixed across all flywheel cycles.
Concretely, the training set comprises the $\approx 200$ rollouts that make up the $100$ preference pairs, each labeled with its VLM-selected dense reward; from each rollout we uniformly subsample $32$ frames and resize them to the backbone's native input resolution. We attach LoRA adapters of rank $128$ ($\alpha = 256$, dropout $0.05$) and optimize with AdamW (learning rate $1\times10^{-4}$, weight decay $0.01$, cosine decay with a short linear warmup) for $5$ epochs at an effective batch size of $32$.

\section{Retrieval and Advantage Labeling Details}
\label{app:advantage}

Recall from Sec.~\ref{sec:advantage} that the dense reward $R_\theta$ is not directly a learning signal: the same return can reflect excellent behavior in a hard state but mediocre behavior in an easy one, so each chunk must be judged \emph{relative to what is achievable from the state it started in}. Rather than learn a value function for this baseline, we estimate it non-parametrically---retrieving chunks that began in visually similar states and ranking the current chunk against them to assign its advantage token.

\paragraph{Observation encoder.}
We embed every frame of every rollout with a frozen, pre-trained vision encoder
$\phi(\cdot)\in\mathbb{R}^{d}$ and $\ell_2$-normalize the output, so all
embeddings live on the unit sphere $\lVert\phi(\cdot)\rVert_2=1$. By default
$\phi$ is DINOv3 (ViT-S/16, $d{=}384$)~\citep{simeoni2025dinov3}.

\paragraph{Visual similarity and comparison set.}
For a query chunk $\tau_i$, our goal is to gather \emph{peer chunks} from other rollouts that began in a visually similar state, so that their returns form a baseline against which $\tau_i$ can be judged.
We measure state similarity in embedding space: each observation $o$ is encoded by the frozen vision encoder $\phi$ and $\ell_2$-normalized, so that $\mathrm{sim}(o,o')=\langle\phi(o),\phi(o')\rangle$ is the cosine similarity between two states. Let $o_{t_i}$ be the \emph{query observation}---the initial frame of $\tau_i$, taken from rollout $e_i$.
We build the comparison set \emph{one donor rollout at a time}. Within each other rollout $e\neq e_i$, we locate the single frame whose observation is most similar to the query,
\begin{equation}
m_e \;=\; \argmax_{0\,\le\, t\,<\,T_e}\ \mathrm{sim}\big(o^{e}_{t},\, o_{t_i}\big),
\end{equation}
and admit that rollout only if this best match is close enough, $\mathrm{sim}(o^{e}_{m_e},\,o_{t_i})\ge\delta$. Restricting each rollout to its \emph{single} best frame prevents one trajectory---whose temporally adjacent frames are nearly identical---from flooding the set with near-duplicate neighbors. Every admitted rollout then contributes exactly one peer chunk, the sub-trajectory $\tau^{(e)}$ that starts at its matched frame $m_e$:
\begin{equation}
\mathcal{N}(\tau_i) \;=\; \big\{\,\tau^{(e)} \;:\; e \neq e_i,\;
\mathrm{sim}\big(o^{e}_{m_e},\, o_{t_i}\big) \,\ge\, \delta \,\big\}.
\end{equation}
By construction, every chunk in $\mathcal{N}(\tau_i)$ starts from a state visually close to $o_{t_i}$; the chunks therefore differ mainly in \emph{what the policy did afterward}, which is precisely the comparison needed to estimate $\tau_i$'s advantage (next paragraph). The threshold $\delta$ is a single global value, tuned so that each comparison set holds $|\mathcal{N}|$ neighbors on average.
Figure~\ref{fig:retrieval_vis} shows, for each task, a query observation alongside the most similar observations retrieved by this procedure.

\begin{figure}[t]
    \centering
    \includegraphics[width=0.72\linewidth]{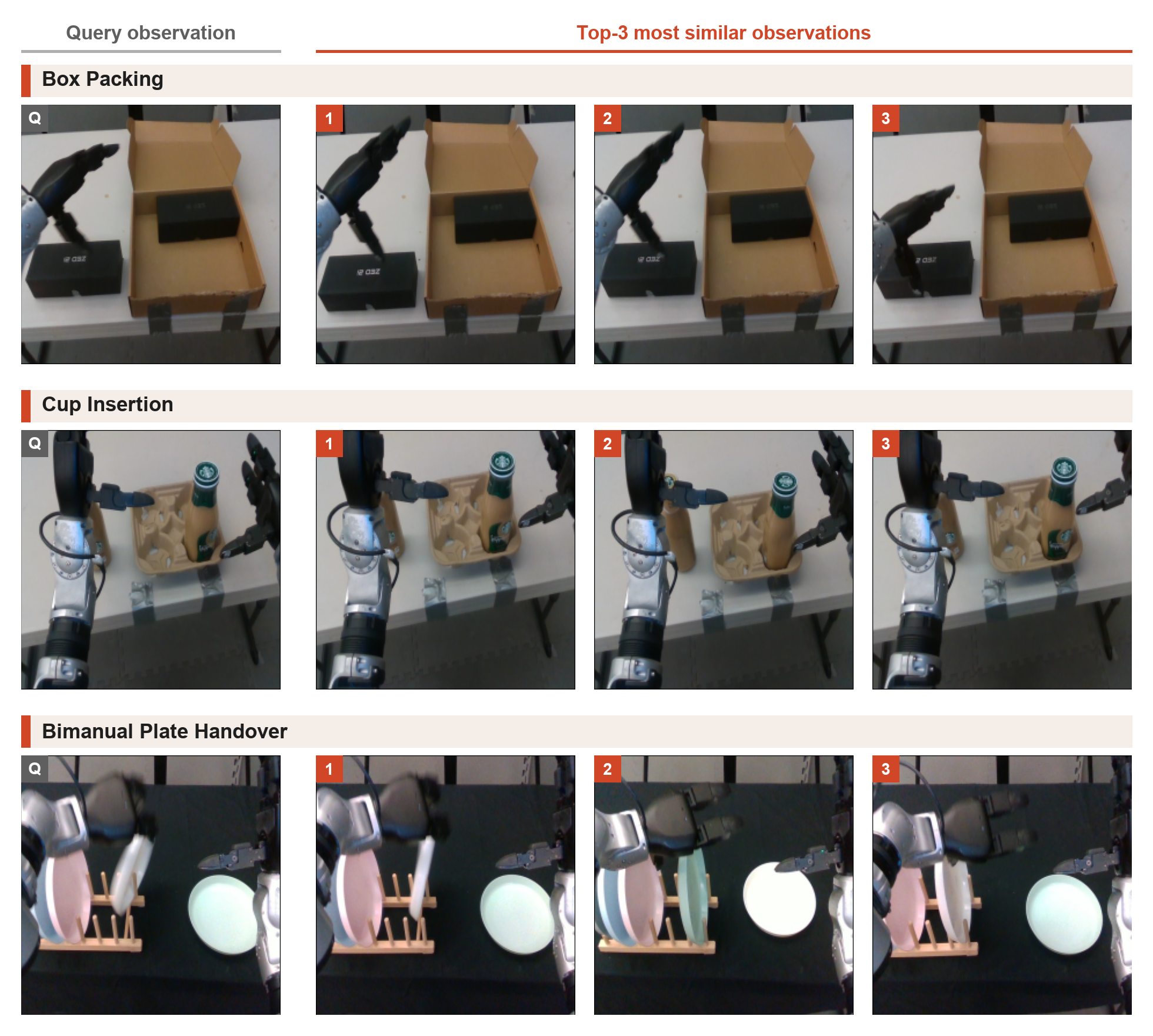}
    \caption{\textbf{Retrieved similar states for advantage estimation.} For each task (rows, top to bottom: Box Packing, Cup Insertion, Bimanual Plate Handover), we show a \emph{query observation} (left) and the \emph{top-3 most similar observations} (right) retrieved from other rollouts via cosine similarity of DINOv3 embeddings. The retrieved states closely match the query in arm pose, object placement, and contact configuration, confirming that the retrieval recovers genuinely comparable states against which chunk returns can be ranked.}
    \label{fig:retrieval_vis}
\end{figure}

\paragraph{Advantage token labeling.}
Given the comparison set $\mathcal{N}(\tau_i)$ built above, we score every chunk by its discounted return over a fixed look-ahead window of $H$ steps from its start,
\begin{equation}
G(\tau_i) \;=\; \sum_{t=t_i}^{t_i+H}\gamma^{\,t-t_i}\, R_\theta(o_t;\ell),
\end{equation}
with discount factor $\gamma$ and a look-ahead horizon $H$ spanning $1.8$ seconds of future observations. We then assign $\tau_i$ a binary advantage token by ranking its return \emph{within} its comparison set: it is labeled \emph{positive} if $G(\tau_i)$ lies in the top $30\%$ of the neighbors' returns and \emph{negative} otherwise,
\begin{equation}
I_i \;=\; \mathbbm{1}\Big(\,G(\tau_i) \,>\, \mathrm{Percentile}_{70}\big\{\,G(\tau_j) \,:\, \tau_j \in \mathcal{N}(\tau_i)\,\big\}\Big).
\end{equation}
Because every chunk in $\mathcal{N}(\tau_i)$ started from a state visually similar to $o_{t_i}$, this percentile threshold adapts to state difficulty: in hard states where most attempts fail, a modest return still earns a positive token, whereas in easy states the bar is correspondingly higher.

\section{Robot Setup}
\label{app:robot_setup}

\paragraph{Hardware.}
All experiments use the Unitree G1 humanoid, which is equipped with two arms, two dexterous multi-fingered hands, and a head carrying two RGB cameras (a regular and a wide-angle lens) that together provide egocentric visual observation of both the workspace and the surrounding scene.

\paragraph{Action space and whole-body control.}
Following~\cite{wei2026psi0}, we adopt a hierarchical whole-body control scheme that decomposes the action space into upper-body and lower-body components. The upper-body action consists of target joint angles for the two arms and hands, while the lower-body action specifies planar linear velocities and a target yaw angle. These high-level commands are tracked by an RL-learned whole-body controller~\cite{xr-teleoperate}, which converts them into the full set of joint-level commands executed on the robot. This decomposition lets the VLA policy issue coarse, task-relevant targets while the low-level controller maintains balance and handles joint-level tracking.

\paragraph{Observation space.}
The policy observation $o_t$ comprises the two head-camera images $(I^{1}_t, I^{2}_t)$---regular and wide-angle---together with the whole-body proprioceptive state $q_t$, which includes the upper-body joint positions and the torso roll, pitch, and yaw as well as the base height.

\paragraph{Policy deployment.}
At each control step the policy outputs an action chunk spanning $1.6$ seconds into the future and replans after the chunk has been executed. All policy variants are deployed through a shared deployment function for fair comparison: at each query we condition the policy on $I = \text{positive}$ and apply classifier-free guidance~\cite{cfgrl} with scale $\beta = 0.2$ to amplify its preference toward high-quality actions. The same deployment configuration---action-chunk length, replanning, conditioning token, and guidance scale---is used for every variant evaluated in this paper, so that performance differences arise solely from the training procedure rather than deployment-time choices.

\section{Evaluations}
\label{app:evaluation}

\subsection{Invasive fine-tuning baseline}
\label{app:invasive}

The $\pi_{0.5}$ action expert is conditioned on a single AdaRMSNorm vector
$\mathbf{c}_{\text{ada}}\in\mathbb{R}^{d}$ produced from the flow-matching timestep and broadcast
to every layer. We compute a per-sample FiLM modulator of $\mathbf{c}_{\text{ada}}$ from
the advantage indicator $I_t$:
\begin{equation}
  [\boldsymbol{\gamma};\boldsymbol{\beta}]
  \;=\; W^{\mathrm{adv}}_{2}\,\mathrm{swish}\!\big(W^{\mathrm{adv}}_{1}\,E_{\mathrm{adv}}[I_t]\big),
  \qquad
  \mathbf{c}_{\text{ada}} \;\leftarrow\; \mathbf{c}_{\text{ada}}\odot(1+\boldsymbol{\gamma})+\boldsymbol{\beta},
  \label{eq:film}
\end{equation}
where $E_{\mathrm{adv}}\!\in\!\mathbb{R}^{3\times d}$ is the indicator embedding vector,
$W^{\mathrm{adv}}_{1}\!\in\!\mathbb{R}^{d\times d}$ and
$W^{\mathrm{adv}}_{2}\!\in\!\mathbb{R}^{d\times 2d}$ are the modulator weights.
Specifically, $W^{\mathrm{adv}}_{2}$ are initialized to zero, giving
$\boldsymbol{\gamma}=\boldsymbol{\beta}=\mathbf{0}$ at step $0$, so $\pi_{0.5}^{\mathrm{Adv}}$ is identical to the loaded $\pi_{0.5}$ checkpoint at the start of every flywheel cycle.
The modulated $\mathbf{c}_{\text{ada}}$ replaces the original at every AdaRMSNorm site of the action expert.

To include the case with no advantage prompt input, we extend $E_{\mathrm{adv}}$ to three rows and route
unlabeled chunks through a learnable \emph{neutral} class:
\begin{equation}
  I_t \;\in\; \{\,0\;\text{(negative)},\;1\;\text{(positive)},\;2\;\text{(neutral)}\,\}.
\end{equation}
All three classes use the same head and produce distinct modulators,
so the model sees a single objective everywhere.
We let the neutral class $I_t{=}2 \ (\text{neutral})$ play the role of
the unconditional branch by giving unlabeled chunks their own
embedding row. Training is therefore a single flow-matching pass with
no indicator dropout and no auxiliary term:
\begin{equation}
  \mathcal{L}(\theta) \;=\;
  \big\|v_\theta(x_\tau,o,\ell,\tau,I_t) - u_\tau\big\|^{2},
  \label{eq:loss}
\end{equation}
where $v_\theta$ is the predicted action-chunk velocity,
$u_\tau$ is the flow-matching target, and $I_t$ is the
per-sample three-class indicator parsed from the prompt suffix. At inference the policy always conditions
on $I_t{=}1 \ (\text{positive})$ with a single suffix pass.
The invasive baseline is run under the same training and iteration setup as the
non-invasive $\pi_{0.5}$+CLIFT entry. Every cycle re-initializes $\pi_{0.5}^{\mathrm{Adv}}$ from the public $\pi_{0.5}$ checkpoint, with the three FiLM modules
initialized from scratch.

\subsection{Evaluation protocol}

\begin{figure}[t]
    \centering
    \includegraphics[width=\linewidth]{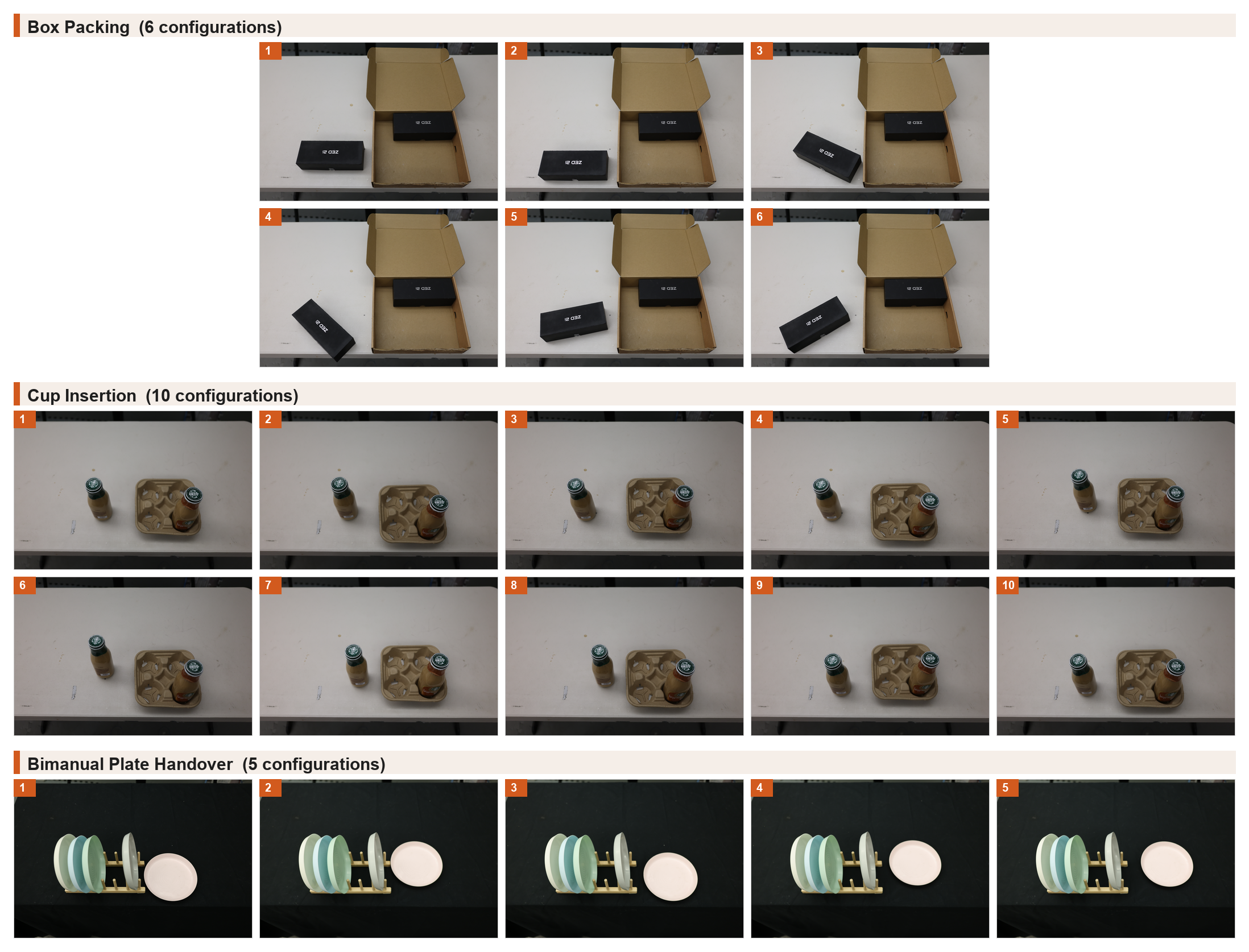}
    \caption{\textbf{Fixed evaluation configurations.} Initial scene layouts used to evaluate every model, variant, and flywheel cycle, varying object positions and orientations. Each task uses its own fixed set: $6$ for Box Packing, $10$ for Cup Insertion, and $5$ for Bimanual Plate Handover. The same configurations are reused identically across all comparisons.}
    \label{fig:eval_configs}
\end{figure}

Evaluating manipulation policies on real hardware is an open problem. Unlike simulation, real rollouts cannot be reset to identical states, and small differences in initial object placement, lighting, or hardware wear can change task difficulty considerably. When different models or methods are each tested under their own loosely controlled conditions, the resulting success rates carry large, uncontrolled variance that can easily swamp the true performance differences between methods.

To control this variance, we fix the evaluation conditions and share them across all comparisons. For each task we define a fixed set of evaluation configurations---$6$ for Box Packing, $10$ for Cup Insertion, and $5$ for Bimanual Plate Handover (Fig.~\ref{fig:eval_configs}). A configuration pins down the initial scene layout---the starting poses of the manipulated objects and their targets---and the configurations are chosen to span the workspace and the range of difficulty the policy must handle. Crucially, the \emph{same} configurations are reused identically for every model, every variant, and every flywheel cycle reported in this paper, so that differences in success rate reflect the policy under test rather than the luck of its initial conditions.

In each flywheel cycle we run $100$ rollouts per task under identical conditions, distributed across that task's configurations, and report the average success rate per task. Before each rollout the scene is reset to its configuration's layout, and the rollout is scored as a binary success or failure against a fixed, task-specific criterion. Because deployment and evaluation coincide in our on-device setting, these same rollouts also serve as the data relabeled for the next flywheel cycle (Sec.~\ref{sec:loop}). All policies are run with the identical deployment configuration, so performance differences arise solely from training rather than deployment-time choices.

\section{Additional Results}
\label{app:additional_result}
\begin{figure}[t]
    \centering
    \includegraphics[width=\linewidth]{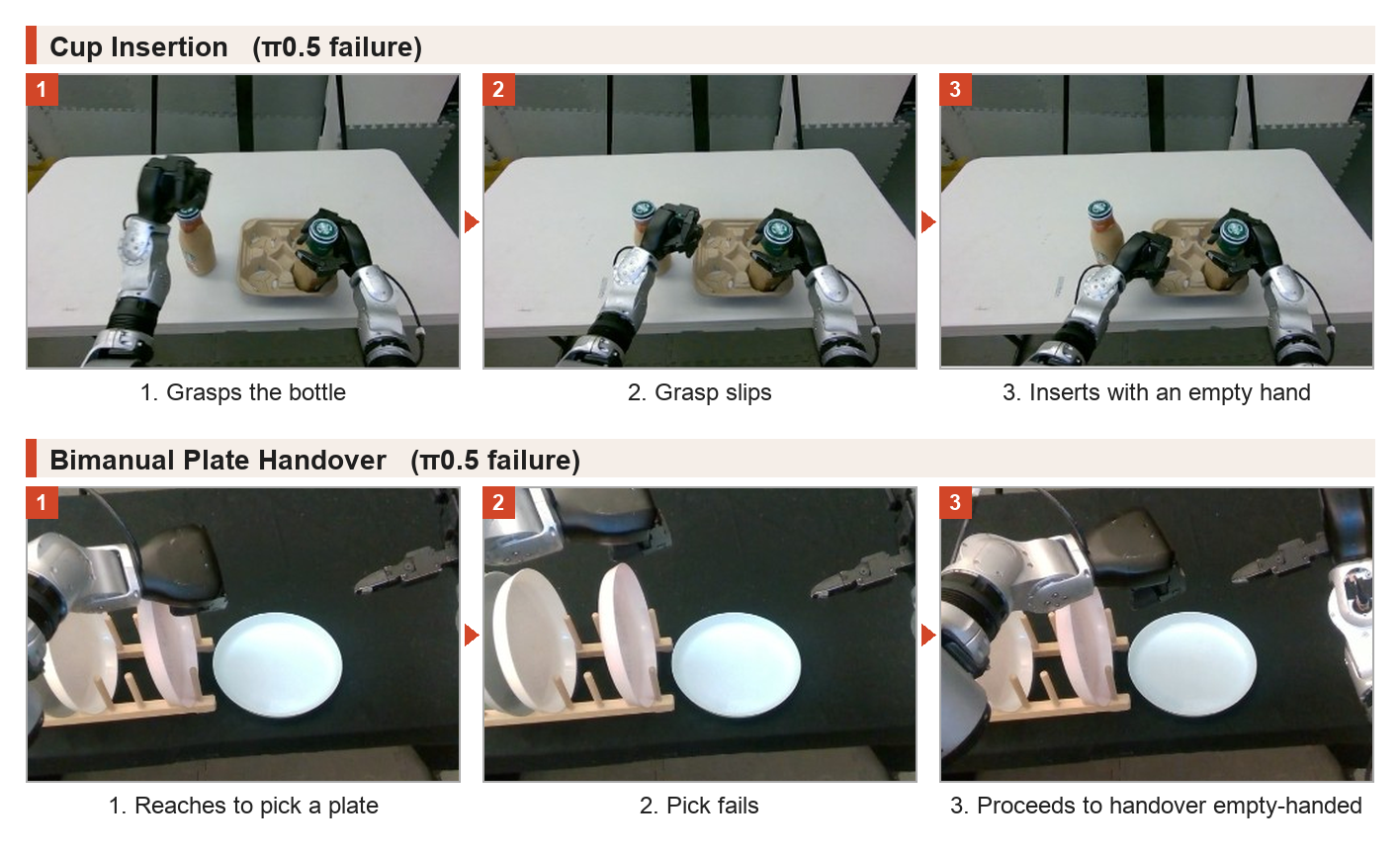}
    \caption{\textbf{Representative $\pi_{0.5}$ failures.} Both tasks exhibit the same failure mode: the policy executes its scripted motion regardless of whether the underlying manipulation actually succeeded. \emph{Top (Cup Insertion):} the policy grasps the bottle (1), the grasp slips and the bottle tips over (2), yet the policy continues the insertion motion with an empty hand (3). \emph{Bottom (Bimanual Plate Handover):} the policy reaches to pick a plate (1), the pick fails (2), yet it proceeds to the handover empty-handed (3). This lack of reactivity to closed-loop feedback is a recurring failure mode for $\pi_{0.5}$.}
    \label{fig:pi05_failure_cases}
\end{figure}

As shown in the main results (Fig.~\ref{fig:results}), applying the identical \oursc pipeline---same reward model, rollout budget, advantage relabeling, and evaluation suite---to the open-weight $\pi_{0.5}$ improves it at every cycle but plateaus at a substantially lower ceiling than GROD, with the gap largest on the most contact-rich tasks. To understand where this gap comes from, we qualitatively examine $\pi_{0.5}$'s rollouts.

\paragraph{$\pi_{0.5}$ is less reactive to closed-loop feedback.}
A recurring failure mode is that $\pi_{0.5}$'s actions do not track what is actually happening in the scene. When an initial grasp slips or misses, the policy frequently proceeds through the rest of the motion as if the grasp had succeeded---lifting and transporting an empty hand---instead of re-attempting the grasp. GROD, by contrast, more often detects the failed grasp and retries before continuing (cf.\ the emergent recovery behavior in Fig.~\ref{fig:qualitative}).

\paragraph{$\pi_{0.5}$ shows weaker emergent behavior.}
Relatedly, $\pi_{0.5}$ exhibits fewer of the corrective and pre-manipulation behaviors that emerge in GROD by the final flywheel cycle (Fig.~\ref{fig:qualitative}). Its gains across cycles come mainly from sharpening behaviors already present in the demonstrations, whereas GROD additionally composes new behaviors---reorienting objects before grasping, retrying after a failed attempt---that extend beyond the demonstration distribution.

\paragraph{Hypothesis: the gap reflects the pretrained prior.}
Because the adaptation procedure is identical for both models, we attribute this gap to the strength of the pretrained prior rather than to the pipeline. Although GROD and $\pi_{0.5}$ share a conceptually similar VLA architecture, GROD is built on a foundation model trained on substantially more and more diverse robot data. We hypothesize that this broader pretraining yields more robust closed-loop perception and reactivity, which in turn lets the same \oursc signal elicit richer corrective behavior and a higher performance ceiling. This is consistent with the controlled comparison in the main text, where even an invasive, full-access adaptation of $\pi_{0.5}$ fails to close the gap to GROD.

}

\end{document}